\documentclass[conference]{IEEEtran}
\IEEEoverridecommandlockouts
\usepackage{cite}
\usepackage{amsmath,amssymb,amsfonts}
\usepackage{algorithmic}
\usepackage{graphicx}
\usepackage{textcomp}
\usepackage{xcolor}
\usepackage{algorithm}
\usepackage{multirow}
\usepackage[utf8]{inputenc}
\usepackage{url}
\usepackage{color}

\def\BibTeX{{\rm B\kern-.05em{\sc i\kern-.025em b}\kern-.08em
    T\kern-.1667em\lower.7ex\hbox{E}\kern-.125emX}}
\begin{document}

\title{A Benchmark dataset for both underwater image enhancement and underwater object detection\\}

\author{\IEEEauthorblockN{Long Chen}
\and
\IEEEauthorblockN{Lei Tong}
\and
\IEEEauthorblockN{Feixiang Zhou}
\and
\IEEEauthorblockN{Zheheng Jiang}
\and
\IEEEauthorblockN{Zhenyang Li}
\and
\IEEEauthorblockN{Jialin Lv}
\and
\IEEEauthorblockN{Junyu Dong}
\and
\IEEEauthorblockN{Huiyu Zhou}
}
\maketitle

\begin{abstract}
Underwater image enhancement is such an important vision task due to its significance in marine engineering and aquatic robot. It is usually work as a pre-processing step to improve the performance of high level vision tasks such as underwater object detection. Even though many previous works show the underwater image enhancement algorithms can boost the detection accuracy of the detectors, no work specially focus on investigating the relationship between these two tasks. This is mainly because existing underwater datasets lack either bounding box annotations or high quality reference images, based on which detection accuracy or image quality assessment metrics are calculated. To investigate how the underwater image enhancement methods influence the following underwater object detection tasks, in this paper, we provide a large-scale underwater object detection dataset with both bounding box annotations and high quality reference images, namely OUC dataset. The OUC dataset provides a platform for researchers to comprehensive study the influence of underwater image enhancement algorithms on the underwater object detection task.
\end{abstract}


\section{Introduction}

During the past few years, underwater object detection (UOD) \cite{b1,b2,b3} has drawn considerable attentions in both marine engineering and aquatic robot. Due to complicated underwater environment and lighting conditions, detecting objects in the water is a challenging problem. The underwater images suffer from serious wavelength-dependent absorption and scattering, which reduces visibility, decrease contrast, and even introduce color casts. This adverse effects limit many practical applications of underwater images and videos in marine biology, archaeology, and ecological. Hence, many underwater image enhancement (UIE) algorithms are employed as a preprocessing step for UOD tasks to improve the detection accuracy of the detectors by boosting the quality of underwater images \cite{b4,b5,b6}. 

\begin{figure*}[h]
\centering
\includegraphics[height=5cm, width=16cm]{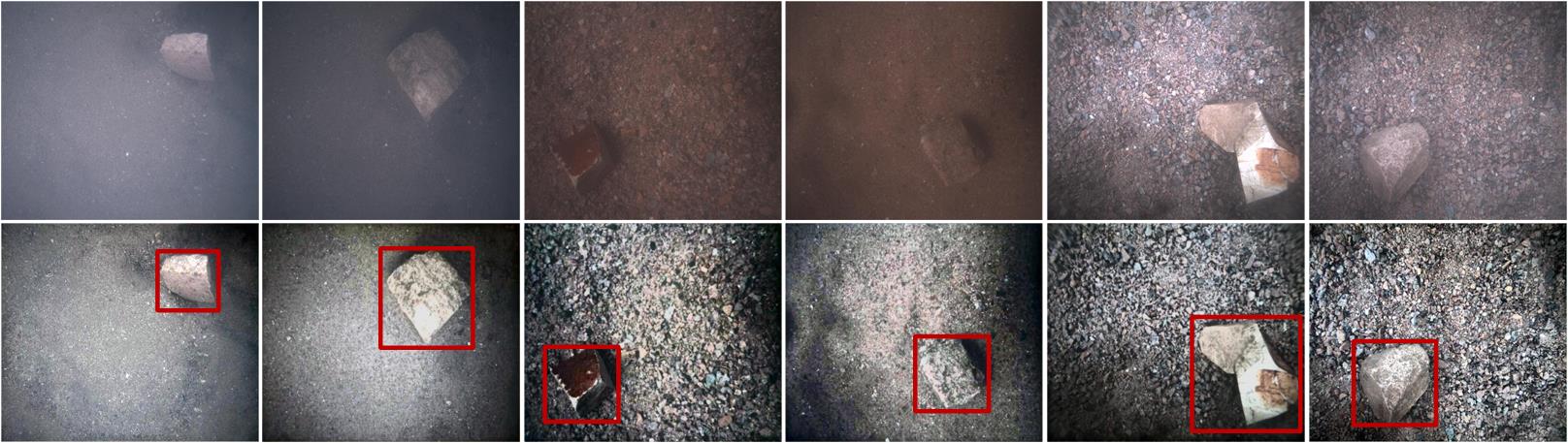}
\caption{Sampling images from the constructed OUC. Top row: raw underwater images taken in diverse underwater scenes; Bottom row: the corresponding reference images and bounding box annotations.}
\label{fig:OUCUIEcases}
\end{figure*}
Despite the prolific work, comprehensive study and insightful analysis of the relationship between UIE and UOD tasks remain insufficient due to the lack of a publicly available underwater image datasets with both bounding box annotations and reference images (i.e, the underwater images without degradation). Since there is no reference images, previous work \cite{b4} only investigated how UIE algorithms influence UOD tasks by study the relationships between the non-reference image quality assessment metrics \cite{b7, b8} and the detection accuracy. However, non-reference image quality evaluation metrics can only explain part characteristics of the image quality and are not always consistent with the human subjective perception \cite{b4}. A comprehensive investigation of the relationship between two tasks should also on the relationship between the detection accuracy and full-reference image evaluation metrics \cite{b9, b10}, which can extensively evaluate the characteristics of image quality in terms of colors, textures, image contents and structures. However, the reference images are necessary when conducts full-reference image quality evaluations. Recently, several underwater image synthesis (UIS) algorithms \cite{b11, b12, b13} had been proposed to synthesize underwater images from high-quality in-air images, then another UIE model is trained on the image pairs to improve the visibility of underwater images. However, the synthetic images are not realistic enough and greatly influence the performance of late UIE models. Differently, Li et al. \cite{b14} employed eleven different UIE algorithms to enhance the underwater images, and choose the high quality reference images from the eleven enhanced results using human subjective perception, i.e, the perception of human. Nevertheless, the subjective perception can be ambiguous and tendentious since different people may have different preferences and biases. Also, human perception is unable to perceive minor differences existing in two visual-similar images. To make up the deficiency of subjective perception, we combine it with the objective assessment to select high quality reference images, which is more robust and dependable than only subjective perception. 

In this paper, we construct an underwater dataset, namely OUC dataset, which  contains underwater images, corresponding reference images and bounding box annotations. To generate robust reference images we propose a novel hybrid reference images generation method which combines subjective perception and objective assessment. Fig.~\ref{fig:OUCUIEcases} presents several sampling underwater images and corresponding reference images with bounding box annotations generated by our hybrid reference image generation method. The raw underwater images in OUC dataset suffer from diverse degrees of haze and contrast decrease. In contrast, the corresponding reference images are characterized by natural color, improved visibility and appropriate brightness. With this dataset, we conduct a comprehensive study of the state-of-the art UIE and UOD algorithms qualitatively and quantitatively. Most importantly, we can investigate how UIE algorithms influence UOD tasks that enables insights into their performance and sheds light on the future research. The main contributions of this paper are summarized as follows: 
\begin{itemize} 
\item We propose a novel reference images generation method which integrates both subjective perception and objective assessment. Through generating dependable high quality reference images for underwater images, we construct a large-scale underwater dataset, namely OUC, which provides underwater images, corresponding high quality reference images, and object-level bounding box annotations.
\item We conduct comprehensive study of the strengths and limitations of different UIE algorithms on the constructed OUC dataset. In addition, this dataset also provide a platform to study the influence of UIE algorithms on the UOD algorithms.
\end{itemize}

\section{Related Work}
\label{sec:rela}
\subsection{Underwater image enhancement}
Underwater images enhancement plays an important role in practical applications that explore and develop the underwater world, such as autonomous underwater vehicles (AUVs) \cite{b15, b16, b17}, unmanned underwater vehicles (UUVs) \cite{b18}, and remotely operated vehicles (ROVs) \cite{b19} navigation. A variety of UIE methods have been proposed and can be divided into three categories. The first line of research is to modify the image pixel values to improve image the contrast, remove haze and correct color casts. It can be divided into spatial domain adjustment and transform domain adjustment. The spatial domain methods \cite{b20, b21, b22} perform adjustment directly in captured underwater images. The transform domain methods \cite{b23} first transform the captured underwater image into a specific domain, and then perform adjustment for haze removal and color correction. These methods can improve the visual quality to some extent, but may degrade details, accentuate noise, introduce artifacts, and cause color distortions.

The second line is physical model-based methods \cite{b24, b25, b26, b27, b28, b29, b30}, which takes the underwater image enhancement as the inverse problem of underwater image degradation. It first construct and estimate a physical image degradation process, then recover the potential high quality image from the estimated physical degradation model. To estimate the parameters of the underwater image degradation model, many UIE algorithms \cite{b25, b26} adapt the classic dark channel prior (DCP) \cite{b24}, which is designed for dehazing in the natural scenes, to underwater scenes. However, these priors do not always works in some cases, e.g., for underwater images that contain white objects or regions, the DCP-based UIE algorithms show limited improvement of the visual quality, or even aggravate the degradation. 

The third line is the deep learning based UIE algorithms, which can be trained using underwater and corresponding reference images. Due to the lack of training pairs, Li et al. \cite{b12} propose an underwater image synthesis model, called WaterGAN, to convert high-quality in-air images and corresponding depth images into underwater-like images. Then, these synthetic image pairs in turn are used to train another two stage deep UIE network. Inspired by Cycle-Consistent Adversarial Networks \cite{b32} which allows learning the mutual mappings between two domains from unpaired data, Fabbri et al. \cite{b33} propose a weakly supervised underwater image synthesis model to synthesize underwater images from high quality in-air images, and then use these synthetic image pairs for training another deep UIE network. Differently, Li et al. [b11] generates training data by exploiting a physically underwater image degradation model and a fixed set of predefined parameters. However, the performances of deep UIE network heavily depend on the quality of the synthetic images, which cannot be perfectly solved by previous underwater image synthesis methods. Therefore, the performance of deep learning-based UIE methods still lag behind conventional state-of-the-art UIE algorithms. To achieve dependable high quality training data, Li et al. \cite{b14} collect a real-world underwater dataset and process the dataset using eleven image enhancement methods. Then, they invite volunteers to select satisfactory reference images via conducting pairwise comparisons. This method generates, at least to some extent, trustworthy reference images by applying subjective perception of human visual system.

\subsection{Underwater Image Quality Evaluation}
Image quality assessment techniques play an important role in underwater image enhancement task, especially beneficial for the development UIE algorithms, it can be divide into subjective assessment and objective assessment. The subjective assessment is usually regarded as the most reliable method of quantifying perceptual quality of content since in most cases such content is meant to be viewed by humans \cite{b34, b35}. However, the subjective assessment depending on the judgement of human observers can be ambiguous and tendentious since the subjective perceptions of different observers are inconsistent. 

The objective image quality assessment metrics are used to measure some important characteristics of the images using statistical numbers, it can be further divided into full-reference image quality assessment metrics \cite{b36} and non-reference image quality assessment metrics \cite{b37, b38}. Most of previous works \cite{b25, b26, b27, b28} only use the non-reference metrics to evaluate UIE algorithms since the underwater datasets do not contain the reference images. Underwater color image quality evaluation metric (UCIQE) \cite{b37} and underwater image quality measure (UIQM) \cite{b38} are two widely used non-reference metrics. UCIQE quantifies the non-uniform color casts, blurring, and low contrast, and then combines these three components in a linear manner. UIQM consists of three attribute measures: a colorfulness measure, a sharpness measure, and a contrast measure. Full-reference metrics are commonly used in cases where reference images exist. For example, the Peak Signal to Noise Ratio (PSNR) is used to measure the similarity between the enhanced underwater images and the reference images in terms of content, and the SSIM \cite{b36} is employed to measure the structure and texture similarity of the enhanced images and the reference images. One major limitation of contemporary objective assessment metrics is that they are usually sensitive to only one or limited types of distortions, while ignoring evaluating distortions of other types, e.g. color distortion, blurry appearance, or decreasing contrast on the underwater images. Therefore, tremendous efforts are highly demanded to more effective image quality assessment methods.

In this section, we construct a large-scale underwater dataset called OUC, which provides underwater images, corresponding reference images and bounding box annotations. We first introduce the collection of the underwater images, then present a novel method to produce the reference images by combining subjective perception and objective assessment.
\begin{figure}[h]
\centering
\includegraphics[height=4cm, width=9cm]{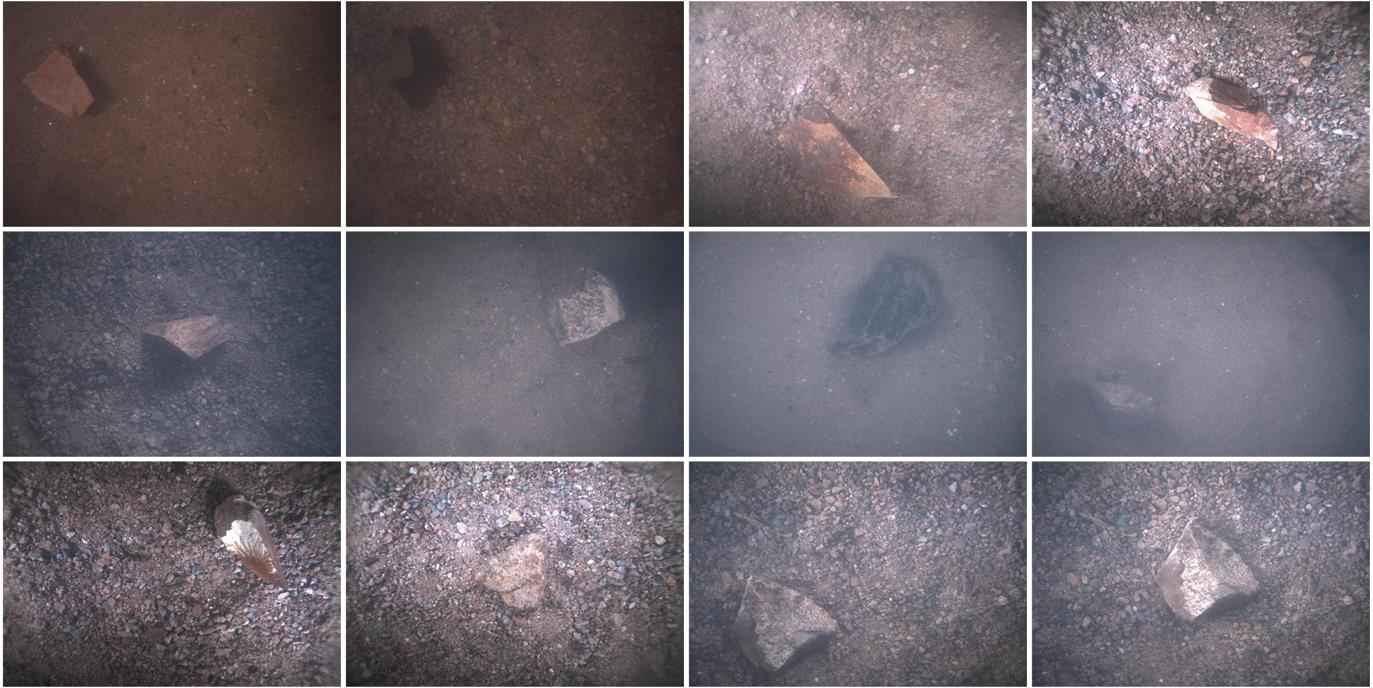}
\caption{Examples of the raw underwater images in the OUC-VISION dataset. These images have different illuminations and haze degrees since they are taken under different underwater environments.}
\label{fig:underwatersample}
\end{figure}
\section{Reference Image Generation}
\label{sec:net}
\subsection{Collection of Underwater Images}
We aims to construct a large-scale underwater dataset which enable researchers to investigate how UIE influence UOD. Hence, the underwater dataset should contain underwater images, reference images and bounding box annotations. When constructing the underwater dataset, we have three objectives:\\
1) The amount of underwater images should be large enough and bounding box level annotations are needed.\\
2) The underwater images should suffer from a diversity of degradation.\\
3) The quality of the reference images should be assured so that the image pairs enable fair evaluation of different UIE algorithms.

To achieve the first two objectives, we choose a large underwater dataset OUC-VISION \cite{b39} that provides underwater images and bounding box annotations. This dataset contains 4,400 underwater images that are captured under different illuminations simulated by a special designed lighting system. In addition, three degrees of turbidity variations, i.e., limpidity, medium and turbidity, are simulated by adding soil to the water. Hence, the underwater images of OUC-VISION suffer from a diversity of illumination variations and turbidity variations. The images are of resolution 486x648 pixels. Fig.~\ref{fig:underwatersample} shows some examples of the raw underwater images in  our OUC dataset which are selected from the OUC-VISION dataset. These images have different characteristics of underwater image (e.g., different color casts, decreased contrast, and haze levels). To achieve the truth-worthy reference images, we propose a novel hybrid reference image generation method which incorporates both subjective perception and objective assessment.
\begin{figure}[h]
\centering
\includegraphics[height=5cm, width=9cm]{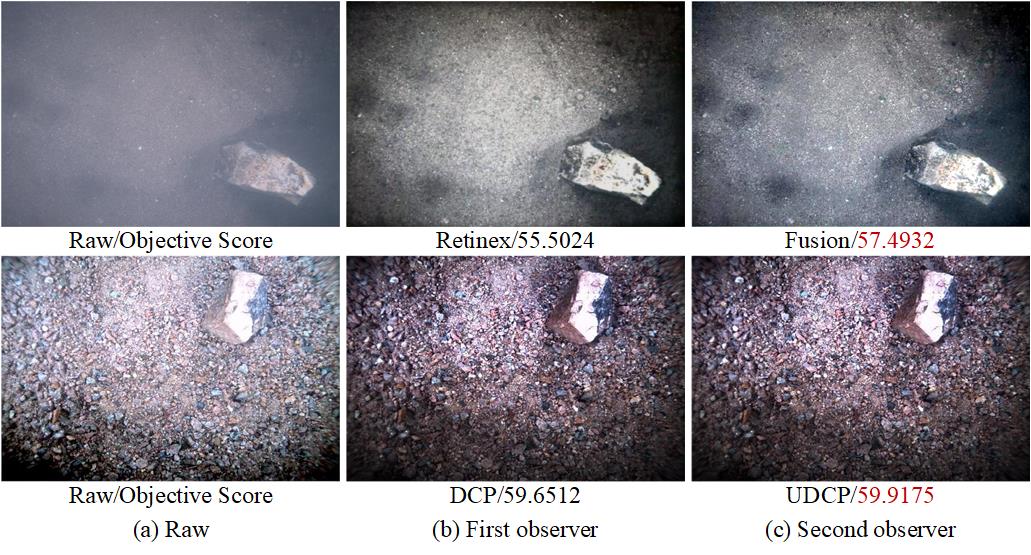}
\caption{The inconsistency of different observers' subjective perception.}
\label{fig:subjectiveerror}
\end{figure}
\begin{figure*}[h]
\centering
\includegraphics[height=8cm, width=16.5cm]{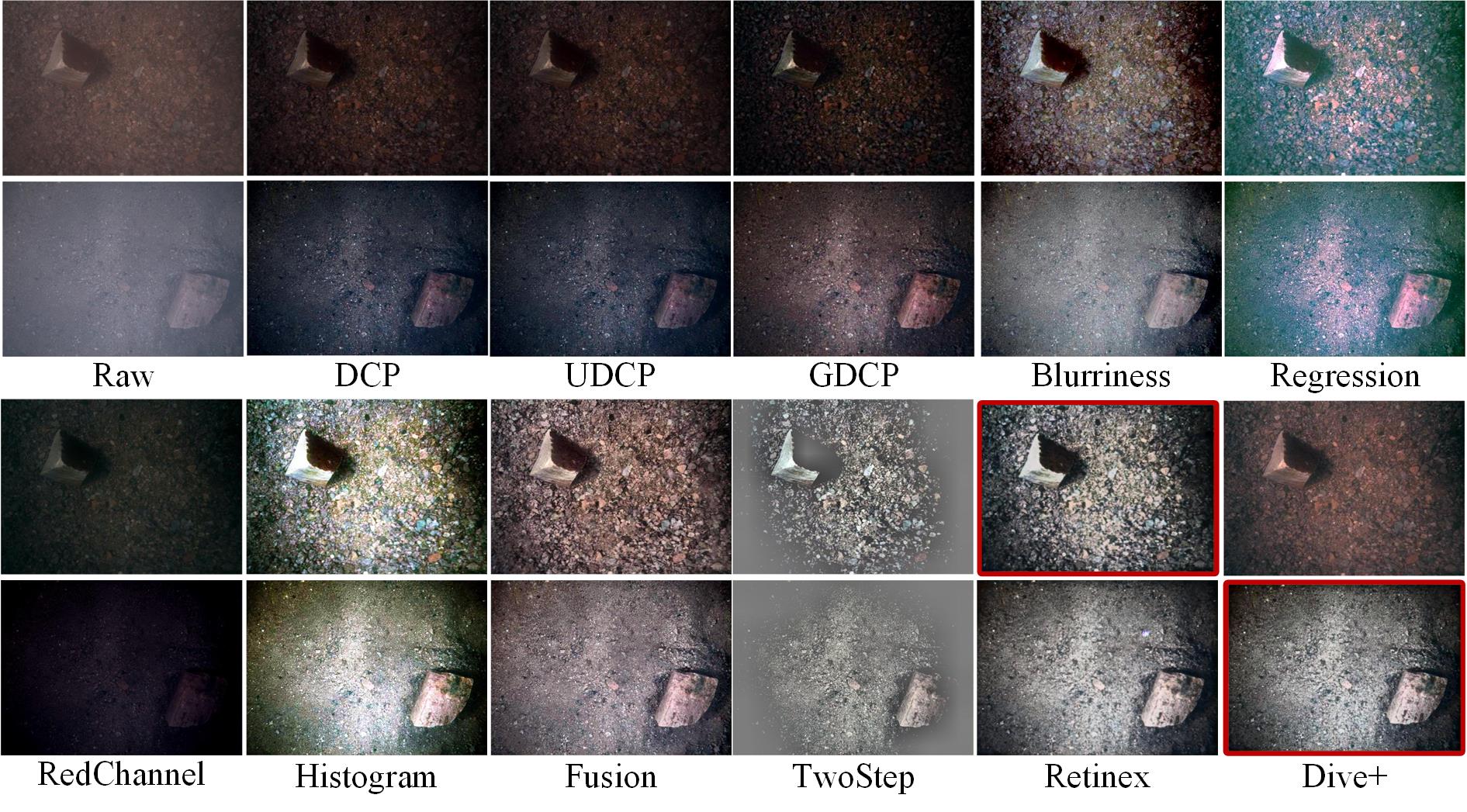}
\caption{Results generated by different methods. From left to right are raw underwater images, and the results of DCP, UDCP, GDCP, Blurriness, Regression, Redchannel, Histogram, Fusion, Twostep, Retinex, and Dive+. Red boxes indicate the final reference images.}
\label{fig:OUCUIEcases}
\end{figure*}
\subsection{Hybrid Image Generation Method}
Previous work \cite{b4} first enhanced underwater images using different UIE algorithms, and then invites multiple observers to select high-quality reference images from the enhanced results, however, using only subjective perception to select images can be tendentious: 1) In many practical cases, the compared images appear the same visual quality that the observers have difficulties in distinguishing them and choosing the best one. For instance, as shown in the top row of Fig.~\ref{fig:subjectiveerror}, two observers select the results of different UIE methods as the final reference images since the visual appearance of two results are extremely similar. 2) The subjective perception is related to Human Visual System, different observers may have different preferences and biases, and no universal standards exist. As shown in the bottom row of Fig.~\ref{fig:subjectiveerror}, the two observers have different preferences and choose different enhanced images as the reference images. To solve these concerns, we propose hybrid reference images generation method by combining the subjective perception and a novel pair-wise objective assessment metric.\\
\textbf{The pairwise objective assessment metric.} 
Specially, when the observes cannot make the decision according to their subjective perceptions in the pairwise comparison, a novel designed pairwise objective assessment metric is employed to select the better one from the two enhanced results. The pairwise objective assessment metric (denotes as $POScore$) depends on the union scores of UIQM and UCIQE, and the pairwise objective score of the $i$-th UIE method's result is formulated as Eq.~\ref{eq:oscore}.\\
\begin{equation}
	POScore_i = UCIQE_i+normUIQM_i, i=1, 2
\label{eq:oscore}
\end{equation} 
For the novel objective metric, we assume UIQM and UCIQE as equally important, so we first normalize UIQM as $normUIQM$. 
\begin{equation}
	normUIQM_1 = \frac{UCIQE_1+UCIQE_2}{UIQM1+UIQM2}*UIQM_1
\end{equation} 
\begin{equation}
	normUIQM_2 = \frac{UCIQE_1+UCIQE_2}{UIQM1+UIQM2}*UIQM_2
\end{equation} 

\begin{table*}[h]
\begin{center}
\renewcommand\tabcolsep{3.5pt}
\caption{Percentage of the reference images from the results of different methods.}
\label{table:percentage}
\begin{tabular}{cccccccccccccc}
\hline
Method & DCP & UDCP & GDCP & Blurriness & Regression & RC & Histogram & Fusion & TwoStep & Retinex & Dive+\\
\hline
Percentage (\%) & 3.68 & 4.50 & 1.30 & 4.40 & 0.00 & 0.00 & 0.70 & 25.10 & 0.00 & 41.72 & 18.60\\
\hline
\end{tabular}
\end{center}
\end{table*}
\textbf{The process of the reference images generation.} 
We first enhance the underwater images using eleven image enhancement methods, including 7 physical-model-based UIE methods (i.e., DCP \cite{b24}, UDCP \cite{b25}, GDCP \cite{b26}, Blurriness \cite{b27}, Regression \cite{b28}, RedChannel \cite{b29} and Histogram \cite{b30}), 3 model-free UIE methods (i.e., Fusion \cite{b20}, Twostep \cite{b21}, and Retinex \cite{b22}), and 1 commercial application for enhancing underwater images (i.e., dive+). We do not exploit deep learning-based UIE methods since we have no training image pairs. At last, we totally obtain 11x4400 enhanced results. With the raw underwater images and the enhanced results, we invite 28 observers, all of whom are students with image processing and computer vision experience, to perform pairwise comparison. They are allowed to draw support from the pairwise objective assessment metric when they cannot make the decision on two ambiguous images in the pairwise comparison. There is no time constraint for observers and zoom-in operation is allowed.

\begin{figure*}[h]
\centering
\includegraphics[height=10cm, width=14cm]{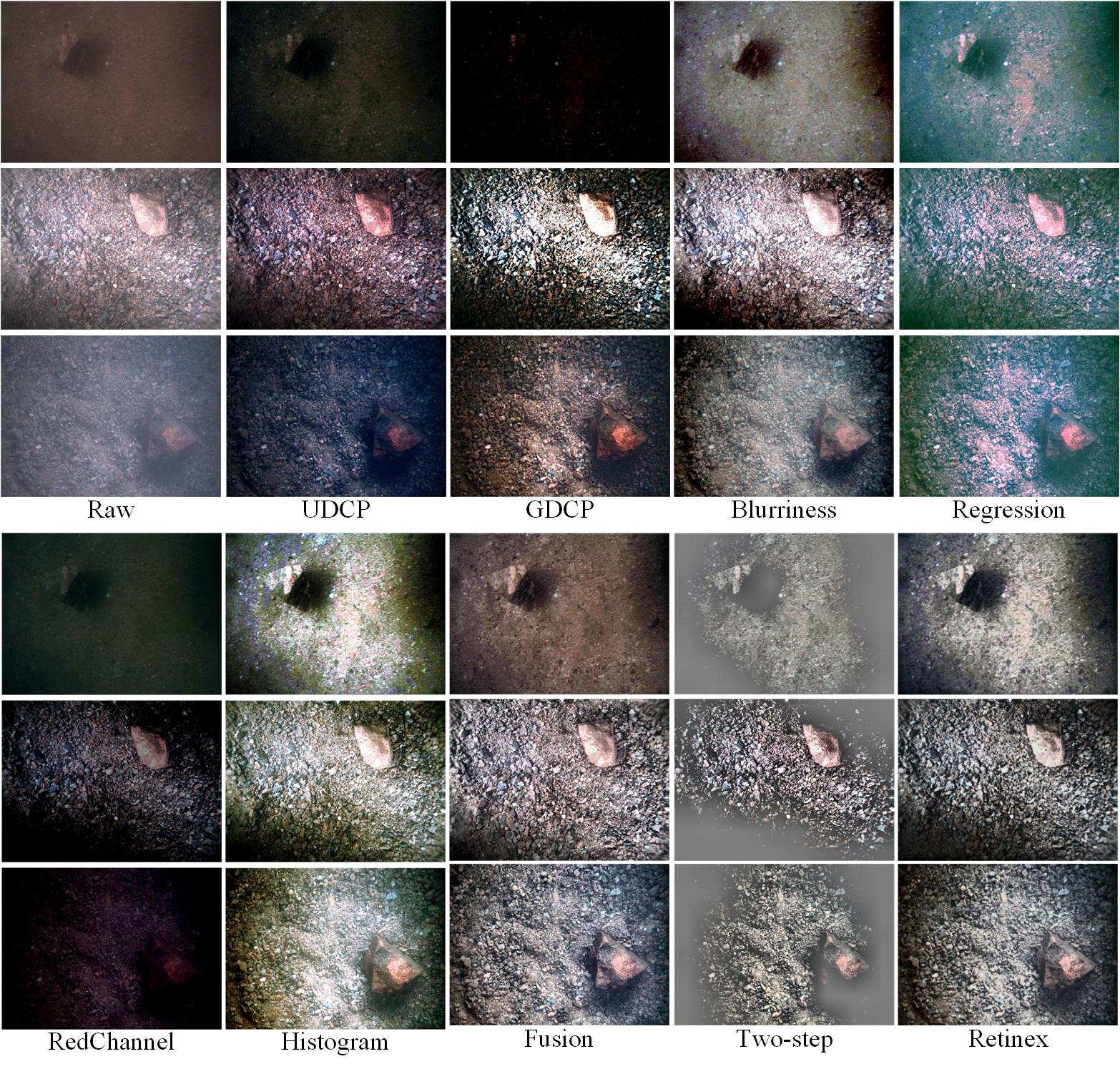}
\caption{Quality comparison of different UIE methods on images captured with different light conditions on the OUC dataset. The top image is captured under light condition 1 and suffer from serious reddish color distortion, the middle one incorporate slightly color distortion and haze effect due to the light condition 2, and the bottom one incorporate evident haze effect and minor color distortion because of light condition 3.}
\label{fig:sot_enhance_OUC_light}
\end{figure*}
The generation of the reference images can be divided into three stages: 1) The reference image selection by a single observer; 2) Check the reference images again and remove unsatisfactory images; 3) Combine the results of all the observers to get the final reference images. For each raw underwater image, the observer is shown two randomly selected enhanced result for pairwise comparison at one time. The observer needs to choose the preferred one or press the button which helps to select the better image using the pairwise objective metric. The result winning the pairwise comparison will be compared again in the next round, until the best one is selected. After the observer finishes the selection work, he/she needs to inspect the reference images set again and remove unsatisfactory images. Then, the reference images of all observers are combined together. For each raw underwater image, if more than half the number of observers remove its corresponding reference images, this underwater image and its reference images will be removed from the final dataset. Finally, the enhanced image selected by more than 50 percentage of observers is selected as the final reference image.  

We totally achieve 3,698 available reference images which have higher quality than the result of any individual UIE methods. To visualize the process of reference image generation, we present some cases that the results of some methods are shown and indicate which one is the final reference image in Fig.~\ref{fig:OUCUIEcases}. Furthermore, the percentage of the reference images from the results of different methods is present in Table~\ref{table:percentage}, the top 1 performance in red, whereas the second top one is in blue.
\begin{table*}[h]
\begin{center}
\caption{Full-Reference image quality and detection accuracy evaluations of different UIE algorithms on the OUC dataset.}
\label{table:quantiOUC_sot}
\resizebox{\textwidth}{12mm}{
\begin{tabular}{cccccccccc}
\hline\noalign{\smallskip}
Methods & UDCP & GDCP & Blurriness & Regression & RedChannel & Histogram & Fusion & Twostep & Retinex\\
\noalign{\smallskip}
\hline
\noalign{\smallskip}
MSE & 3.6147 & 2.4115 & 0.6100 & 0.4294 & 7.1857 & 0.5325 & \textbf{0.2816} & 1.6242 & 0.3469\\
PSNR & 12.9696 & 15.7764 & 20.9975 & 22.1125 & 9.7217 & 21.3031 & \textbf{28.5319} & 16.1558 & 28.0519\\
SSIM & 0.4807 & 0.6316 & 0.7239 & 0.5343 & 0.1798 & 0.7531 & 0.8794 & 0.6108 & \textbf{0.8886}\\
PCQI & 0.4181 & 0.5660 & 0.6493 & 0.6620 & 0.1694 & 0.8089 & \textbf{0.8940} & 0.4785 & 0.8367\\
mAP & 87.1 & 86.9 & 86.4 &  81.6 & 41.6 & 81.5 & 83.9 & 74.8 & \textbf{87.2}\\
\hline
\end{tabular}}
\end{center}
\end{table*}
\subsection{Evaluation of different UIE algorithms on OUC dataset.}
We also evaluate different UIE algorithms on the OUC dataset. We resize all the images into 512x512 pixels, and divide the OUC dataset into training set containing 2,500 image pairs and testing set containing 1,198 image pairs.  Fig.~\ref{fig:sot_enhance_OUC_light} shows qualitative comparisons of different UIE algorithms on underwater images different light conditions. The top image is captured under light condition 1 and suffer from serious reddish color distortion, the middle one incorporate slightly color distortion and haze effect due to the light condition 2, and the bottom one incorporate evident haze effect and minor color distortion because of light condition 3. We observe that none of the physical model-based methods were able to solve the reddish color distortion. This is because the existence of reddish underwater images has violated the physical prior. In water, the red light first disappears because of its longest wavelength, followed by the green light and then the blue light. Such selective attenuation in water results in the greenish and bluish underwater images, and seldom reddish underwater images. In addition, among all the physical model-based algorithms, Regression, Histogram and RedChannel cannot well deal with underwater images with all kinds of light conditions. Regression introduces serious blueish color distortion due to its inaccurate color correction algorithm, and Histogram introduces greenish color distortion due to its histogram distribution prior. RedChannel greatly decreases the brightness which seriously smears the details of images. Moreover, Two-step, one of the non-physical model-based algorithms, also fails on all kinds of light conditions. It over-enhances the contrast and generates unnatural images. In contrast, OurPatch well deals with all kinds of underwater images in terms of both color distortion and haze effect, while the remainders only work in special scene. For example, GDCP and Fusion remove the haze effects and greatly improve the visibility of underwater images captured under the light conditions 2 and 3. UDCP greatly removes haze, however, it introduces bluish color tone into the images captured under the light condition 2 and reddish color tone into the images captured under the light condition 3. Blurriness greatly remove haze on images captured under all of the three light conditions, but fails to remit the color casts in images captured under the light condition 1. These physical model-based methods all fail on some underwater images captured under specific light conditions due to the limitations of the priors used in them. Among the non-physical model-based methods, Retinex greatly remove haze and remit color cast in all kinds of underwater images, but its results suffer from limited saturation.

Table~\ref{table:quantiOUC_sot} reports the quantitative scores of different UIE algorithms on the testing set of OUC. In terms of the four full-reference image quality metrics, Fusion achieves the best MSE, PSNR, and PCQI scores, while Retinex achieves the best SSIM score. In terms of mAP, Retinex achieve the best detection accuracy 87.2 mAP.

\section{Conclusion}
In this paper, we propose a novel reference images generation method which integrates both subjective perception and objective assessment. Through generating dependable high quality reference images for underwater images, we construct a large-scale underwater dataset, namely OUC, which provides underwater images, corresponding high quality reference images, and object-level bounding box annotations.

\end{document}